\documentclass{article}

\usepackage{arxiv}

\usepackage[utf8]{inputenc} 
\usepackage[T1]{fontenc}    
\usepackage{hyperref}       
\usepackage{url}            
\usepackage{booktabs}       
\usepackage{amsmath}
\usepackage{amsfonts}       
\usepackage{nicefrac}       
\usepackage{microtype}      
\usepackage{cleveref}       
\usepackage{lipsum}         
\usepackage{graphicx}
\usepackage{natbib}
\usepackage{doi}
\crefname{figure}{Figure}{Figures}
\crefname{table}{Table}{Tables}

\title{Local Energy Distribution  Based Hyperparameter Determination for Stochastic Simulated Annealing}

\author{
	Naoya Onizawa \\
	Research Institute of Electrical Communication\\
	Tohoku University\\
    Sendai, Japan 980-8577 \\
	\texttt{naoya.onizawa.a7@tohoku.ac.jp} \\
	\And
    Kyo Kuroki \\
	Research Institute of Electrical Communication\\
	Tohoku University\\
	Sendai, Japan 980-8577 \\
	\texttt{kyo.kuroki.q7@tohoku.ac.jp} \\
	\And
	Duckgyu Shin \\
	Research Institute of Electrical Communication\\
	Tohoku University\\
	Sendai, Japan 980-8577 \\
	\texttt{duckgyu.shin.p4@dc.tohoku.ac.jp} \\
	\And
	Takahiro Hanyu \\
	Research Institute of Electrical Communication\\
   Tohoku University\\
   Sendai, Japan 980-8577 \\
  \texttt{takahiro.hanyu.c4@tohoku.ac.jp} \\ 
}

\renewcommand{\shorttitle}{Local Energy Distribution  Based Hyperparameter Determination}

\hypersetup{
pdftitle={Local Energy Distribution  Based Hyperparameter Determination for Stochastic Simulated Annealing},
pdfsubject={LED},
pdfauthor={Naoya ~Onizawa},
pdfkeywords={Combinatorial optimization,stochastic computing},
}

\begin{document}
\maketitle

\begin{abstract}
	This paper presents a local energy distribution based hyperparameter determination for stochastic simulated annealing (SSA).
	SSA is capable of solving combinatorial optimization problems faster than typical simulated annealing (SA), but requires a time-consuming hyperparameter search.
	The proposed method determines hyperparameters based on the local energy distributions of spins (probabilistic bits). 
	The spin is a basic computing element of SSA and is graphically connected to other spins with its weights. 
	The distribution of the local energy can be estimated based on the central limit theorem (CLT). 
	The CLT-based normal distribution is used to determine the hyperparameters, which reduces the time complexity for hyperparameter search from $\mathcal{O}(n^3)$ of the conventional method to $\mathcal{O}(1)$.
	The performance of SSA with the determined hyperparameters is evaluated on the Gset and K2000 benchmarks for maximum-cut problems. 
	The results show that the proposed method achieves mean cut values of approximately 98\% of the best-known cut values.
	
\end{abstract}



\section{INTRODUCTION}

Combinatorial optimization  is used to solve many practical problems in various fields and involves finding the optimal solution for a given objective function subject to a set of constraints \cite{QA_review}.
Combinatorial optimization problems are often NP-hard, meaning that finding the optimal solution requires an exponentially large amount of time with respect to the problem size \cite{NP-hard}.
One such approach is simulated annealing, which is a stochastic optimization method inspired by the physical annealing process in materials science \cite{SA1,SA2}.
SA has been successfully applied to various combinatorial optimization problems, such as the traveling salesman problem, the graph coloring problem, and the maximum cut problem \cite{SA_max-cut}.
%
%
Quantum annealing (QA) is another optimization method that uses quantum mechanics to solve combinatorial optimization problems \cite{QA1,QA2}.
It employs quantum devices that exploit quantum superposition and entanglement to explore the search space more efficiently than classical algorithms \cite{QA_Google}; however, QA devices are still in the early stage of development and have limited qubit coherence times and connectivity, which restricts their ability to solve large-scale problems \cite{D-Wave5000}.

Recently, a novel method known as stochastic-computing-based simulated annealing (SSA) has been introduced, which significantly expedites the annealing process compared to traditional SA and QA methods \cite{SSA}.
SSA is a parallel variant of simulated annealing that utilizes an approximation of probabilistic bits (p-bits) \cite{IL}, implemented through stochastic computing \cite{stochastic_first,stochastic}. 
Its flexibility allows for implementation in both software and hardware \cite{JETCAS_SSA}, potentially scaling up to handle large-scale problems.
Combinatorial optimization problems are expressed using an Ising model \cite{Ising}, which comprises spin states denoted as $\sigma$, spin biases symbolized by $h$, and weights between spins represented by $J$. The Ising model embodies an energy.
In SSA, a pseudo inverse temperature is incrementally raised to determine a solution that corresponds to the global minimum energy, using random signals.
To enhance performance (i.e., to increase the likelihood of reaching the global minimum energy), hyperparameters related to the pseudo inverse temperature and the random signals must be fine-tuned. 
This process, however, can be time-consuming due to the significant computational cost involved.

In this paper, we introduce a statistical method for determining hyperparameters for SSA that can eliminate the time-consuming hyperparameter search.
In the proposed method, the hyperparameters are calculated based on the local  energy distributions of spins.
The local energy at each spin is calculated by the spin bias and the summation of the  multiplication of the spin weights and other spin states.
As the number of connections from the other spins is sufficiently large, the distribution of the local energy can be estimated based on the central limit theorem (CLT).
Therefore, the local-energy distribution can be approximated by the normal distribution, which determines the hyperparameters, such as the pseudo inverse temperature and the noise signals.
Compared with a conventional method that uses a grid search (i.e. typical hyperparameter search) in \cite{SSA}, the time complexity for the hyperparameter search is reduced to $\mathcal{O}(1)$  from  $\mathcal{O}(n^3)$.
Using the determined hyperparameters, a unique noise control at each spin is also introduced for SSA to enhance the performance.
The determined hyperparameters are evaluated in maximum-cut (MAX-CUT) problems \cite{max-cut} that are a typical combinatorial optimization  problem.
On the MAX-CUT problem benchmarks, such as Gset \cite{G-set} and K2000 \cite{K2000}, the mean cut values using the proposed method reach approximately 98\% of the best-known cut values for 1,000 cycles.

The contributions of the paper are:
\begin{enumerate}
	\item Introducing a new statistical method for determining hyperparameters in SSA that can eliminate the time-consuming hyperparameter search required by the conventional method.
	\item Demonstrating the effectiveness of the proposed method in solving the MAX-CUT problems and achieving mean cut values that are close to the best-known cut values.
	\item Reducing the time complexity for hyperparameter search from  $\mathcal{O}(n^3)$ to  $\mathcal{O}(1)$ compared to the conventional method.
\end{enumerate}
The rest of the paper is structured as follows. 
\cref{sec:preliminary} reviews SSA and its hyperparameters. 
\cref{sec:hyperparameter} presents the proposed statistical method for determining the hyperparameters in SSA. 
\cref{sec:individual} introduces the unique noise magnitude for each spin. 
\cref{sec:evaluation} evaluates the determined hyperparameters and compares SSA with conventional SA. 
\cref{sec:discussion} discusses the hyperparameter determination with the conventional hyperparameter search.
Finally, \cref{sec:conclusion} concludes the paper.

\section{PRELIMINARY}
\label{sec:preliminary}

\subsection{Stochastic simulated annealing (SSA)}

There are several SA methods, such as serial updating \cite{SA_max-cut} and parallel tempering \cite{Ising_PT}.
Recently, SSA has been presented as one of the SA methods \cite{SSA}.
SSA is designed based on p-bit-based SA (pSA), which was introduced in \cite{pbits_general}.
A p-bit is  a probabilistic bit that can be one of two spin states, `+1' and `-1'.
It  has been proposed for use in invertible logic, an unconventional computing technique \cite{IL,CIL,CIL_training}.
pSA, implemented on an underlying Boltzmann machine \cite{Boltzmann1984}, realizes parallel updating of the spins for fast simulated annealing.
SSA approximates the behavior of p-bit using stochastic computing, which overcomes the  slow convergence to the global minimum energy of pSA.
Compared to other SA methods, SSA achieved the highest maximum-cut value in K2000 \cite{ICECS2022}.

\begin{figure}[t]
	\centering
	\includegraphics[width=1.0\linewidth]{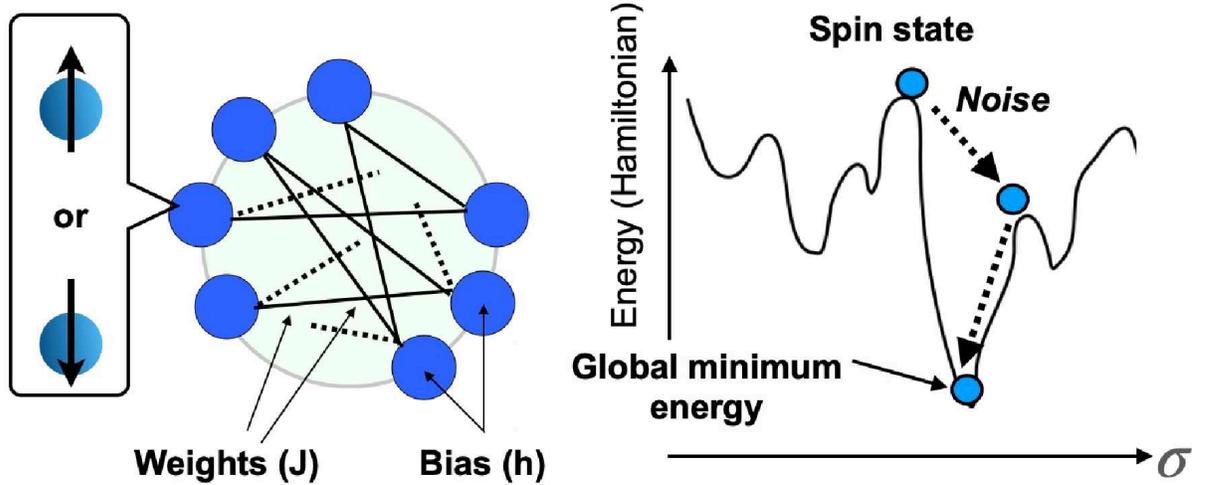}
	\caption{Stochastic simulated annealing  (SSA) \cite{SSA} based on a spin network that consists of  spins, spin biases $h$, and spin weights $J$. Spin states are flipped between '+1' and '-1' by noise signals to reach the global minimum energy of the Hamiltonian.}
	\label{fig:SSA}
\end{figure}

\cref{fig:SSA} illustrates SSA, which is on a spin network designed using spins, spin biases $h$, and spin weights $J$ between spins.
Each spin state $\sigma$ can be either `-1' and `+1'.
The spin network is an Ising model \cite{Ising} that represents a Hamiltonian (energy function) as follows:
\begin{equation}
	H(\sigma) = - \sum_i h_i\sigma_i - \sum_{i < j} J_{ij}\sigma_i\sigma_j.
	\label{eqn:Ising_SA}
\end{equation}
To solve a combinatorial optimization problem, the problem is mapped to the Hamiltonian coefficients of $h$ and $J$. 
The coefficients differ depending on the problem, such as graph isomorphism and MAX-CUT problems \cite{QA_GI,max-cut}. 
During the annealing process, spin states are flipped between ‘+1’ and ‘-1’ in an attempt to reach the global minimum of the Hamiltonian.

In SSA, the spin behavior is modeled by approximating the p-bit using integral stochastic computing (ISC) \cite{SDNN}. 
Note that ISC is an extended version of typical stochastic computing \cite{stochastic_first,stochastic}, which can be used for area-efficient hardware implementation \cite{Sldpc1,Simage,SIIR}. 
An $i$-th spin has a bias $h_i$ and edge weights $J_{ij}$ from/to other spins. 
At each cycle, the spin state is updated as follows:
\begin{subequations}
	\begin{equation}
		I_i(t+1) = h_i+\sum_j J_{ij}\cdot \sigma_j(t) + n_{rnd}\cdot r_i(t),
		\label{eqn:I_SSA}
	\end{equation}
	\begin{equation}
		I{s}_i(t+1)=
		\begin{cases}
			I_0(t)-\alpha, \text{if} \ I{s}_i(t) + I_i(t+1) \geq I_0(t) \\
			-I_0(t), \text{else if} \ I{s}_i(t) + I_i(t+1) < -I_0(t) \\
			I{s}_i(t) + I_i(t+1),  \text{otherwise}
		\end{cases}
		\label{eqn:updown_SSA}
	\end{equation}
	\begin{equation}
		\sigma_i(t+1)=
		\begin{cases}
			1,& \text{if} \ I{s}_i(t+1) \geq 0 \\
			-1, & \text{otherwise},
		\end{cases}
		\label{eqn:m_SSA}
	\end{equation}
	\label{eqn:SSA}
\end{subequations}
where $\sigma_i(t) \in \{-1,1\}$ and $\sigma_i(t+1) \in \{-1,1\}$ is binary input and output spin states, respectively.
$I_i(t+1)$ and  $Is_i(t+1)$ are real-valued internal signals and $n_{rnd}$ is the magnitude of a random signal, $r_i(t) \in \{-1,1\}$.
$I_0$ is the pseudo inverse temperature and $\alpha$ is the minimum resolution of data representation.
If only integer values are used in \cref{eqn:SSA}, $\alpha$ is 1 \cite{SSA}.
If floating-point values are used, $\alpha$ can be approximated by 0.
%

\subsection{Hyperparameter search for SSA}

\begin{figure}[t]
	\centering
	\includegraphics[width=0.6\linewidth]{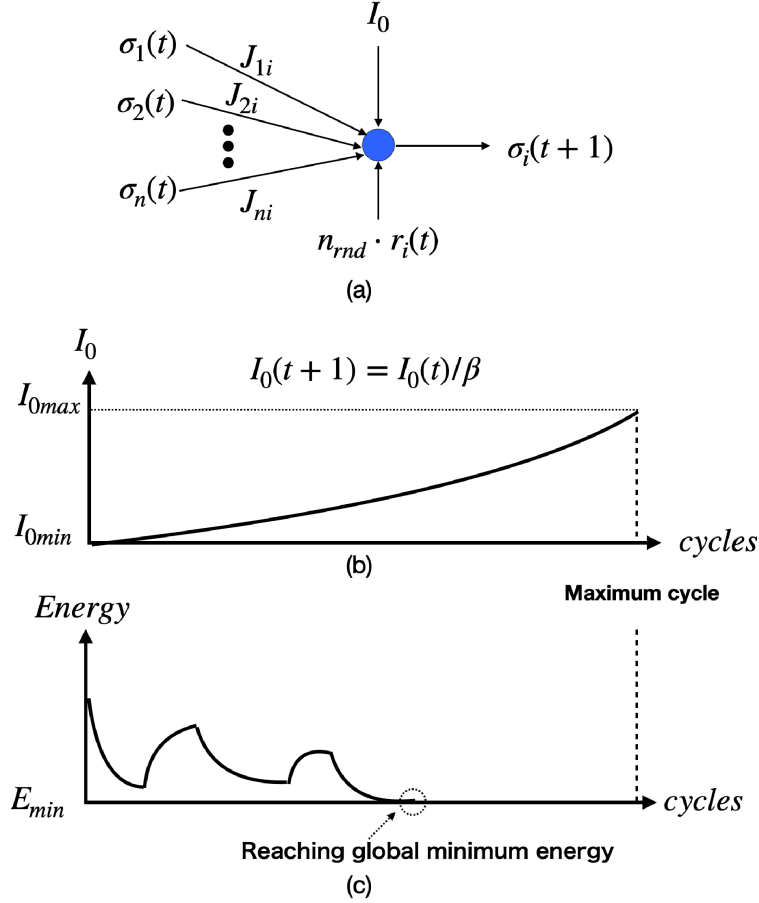}
	\caption{Annealing process of SSA: (a) spin, (b) $I_0$ control, and (c) energy transition. In SSA, $I_0$ is gradually increased from $I_{0min}$ to $I_{0max}$ with weighted noise signals $n_{rnd}\cdot r_i(t)$. These hyperparameters need to be selected for the best performance of SSA.}
	\label{fig:hyperparameter}
\end{figure}

Let us explain the annealing process of SSA using \cref{fig:hyperparameter}. 
At each cycle, $I_0$ is gradually increased from $I_{0min}$ to $I_{0max}$ as $I_0(t+1)=I_0(t)/\beta$.
The parameter $I_0$ controls the strength of the external field that acts on each spin.
When $I_0$ is small, the spin states can be easily flipped to search for many spin states. 
This helps the spin states escape from local energy minima and explore a wider region of the solution space.
When $I_0$ is large, the spin states can be stabilized in an attempt to reach the global minimum energy. 
Additionally, $n_{rnd}$ is an important parameter used to control the stability of the spin states.

\begin{figure}[t]
	\centering
	\includegraphics[width=0.6\linewidth]{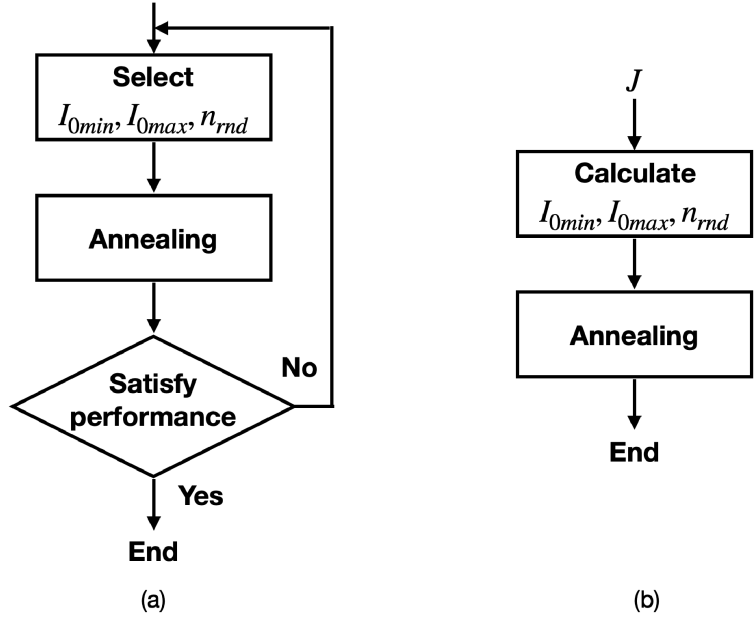}
	\caption{Selecting hyperparameters in SSA: (a) conventional method \cite{SSA,ICECS2022} and (b) proposed method. The proposed method statistically determines hyperparameters without searching while the conventional method uses the grid/random search to find good hyperparameters.}
	\label{fig:flow}
\end{figure}

To achieve a high probability of reaching the global minimum energy, the three hyperparameters of $I_{0min}$, $I_{0max}$ and $n_{rnd}$ need to be carefully selected. 
In previous studies \cite{SSA,ICECS2022}, a grid search or a random search has been used to select these hyperparameters, as shown in \cref{fig:flow} (a). 
In a grid search, a search space is defined as a grid of hyperparameter values, and every position in the grid is evaluated. 
In a random search, a search space is defined as a bounded domain of hyperparameter values, and points are randomly sampled within that domain. 
If good hyperparameters are not found, the search process is iteratively carried out until the required performance (e.g., probability of reaching the global minimum energy) is satisfied.
While these methods are simple, they often require a long time to find good hyperparameters. 
In particular, when solving large combinatorial optimization problems, the time-consuming hyperparameter search could be a critical issue. 
In this paper, we determine the hyperparameters statistically by calculation, without performing a search, while achieving high performance of SSA, as shown in \cref{fig:flow} (b).

\section{HYPERPARAMETER DETERMINATION BASED ON LOCAL ENERGY DISTRIBUTION}
\label{sec:hyperparameter}

\subsection{Local energy distribution}

\begin{figure}[t]
	\centering
	\includegraphics[width=0.6\linewidth]{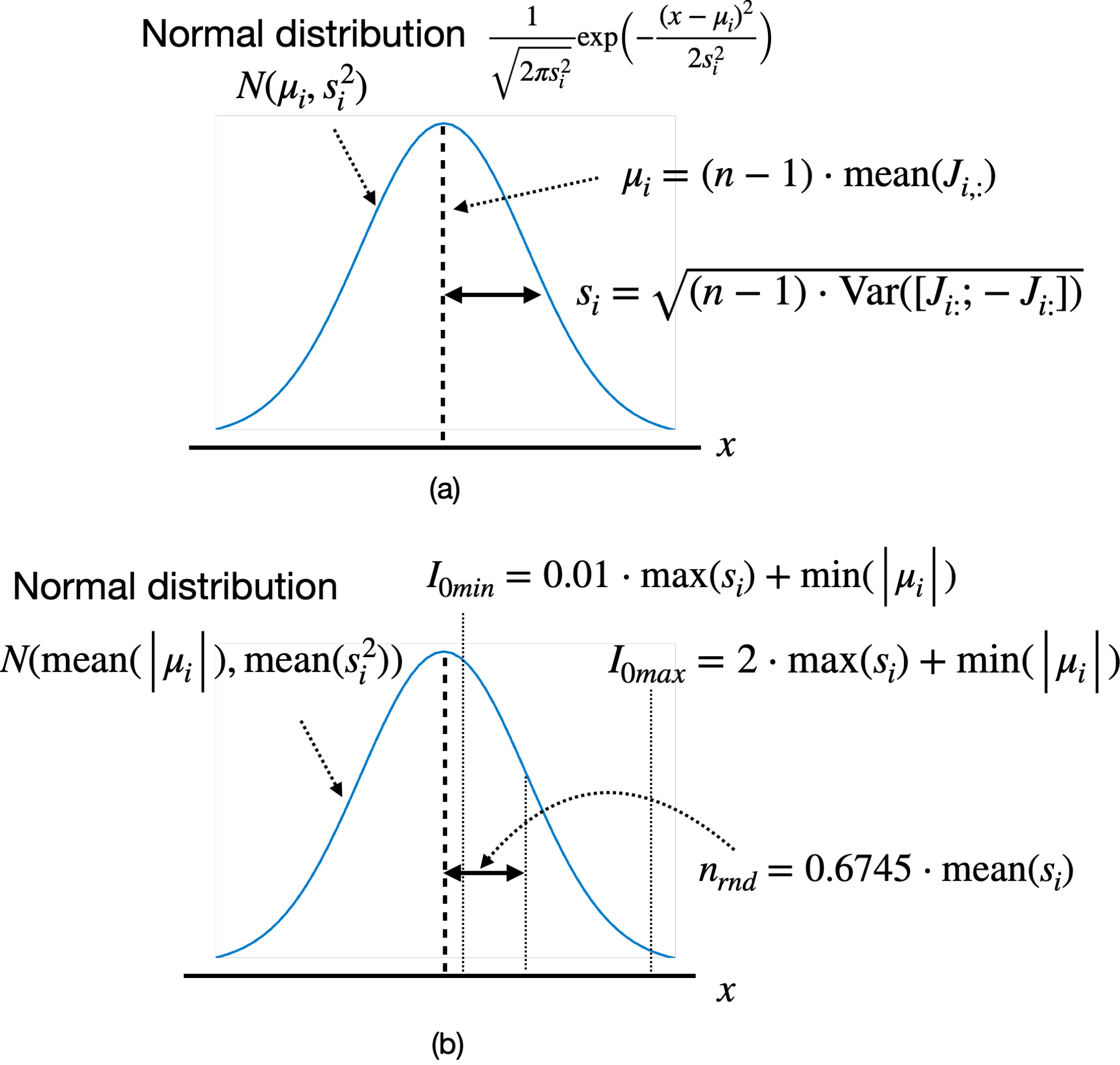}
	\caption{Hyperparameter determination in SSA: (a) local energy distribution at $i$-th spin approximated by the normal distribution and (b) hyperparameter determination using the mean of all the local energy distributions. The three hyperparameters of $I_{0min}$, $I_{0max}$, and $n_{rnd}$ are determined using the normal distribution $N({\rm mean}(\left|\mu_i\right|), {\rm mean}(s_i^2))$.}
	\label{fig:selection}
\end{figure}

In the spin network, the spins interact with each other via pairwise interactions, which are often represented by the edge weights $J_{ij}$ in a graph representation of the Ising model.
The total energy of the spin network is defined as per \cref{eqn:Ising_SA}, whereas the local energy at the $i$-th spin is defined as follows:
\begin{equation}
	H_L(\sigma_l) = -h_i\sigma_i - \sum_{i\neq j}J_{ij}\sigma_j\sigma_i.
	\label{eqn:local}
\end{equation}
%
%
The local energy varies depending on the states of other spins $\sigma_j$ that are connected to it with the edge weights $J_{ij}$. 
As the number of edges connected to a spin becomes larger, the distribution of the local energy becomes more and more Gaussian, which can be explained by the central limit theorem (CLT). 
In this case, the random variables are the edge weights $J_{ij}$, and the sum is over all the edges connected to a spin.

In \cref{fig:selection} (a), the local energy distribution at the spin is illustrated, where $n$ is the total number of spins representing a combinatorial optimization problem. 
The $i$-th spin connects $(n-1)$ edges to other spins with the edge weights $J_{i:}$. 
Note that $J_{i:}$ is a vector containing all the edge weights connected to the $i$-th spin.
Using CLT, the local energy distribution can be approximated by a normal distribution $N(\mu_i,s_i^2)$ with mean $\mu_i$ and standard deviation $s_i$ calculated as follows: 
\begin{subequations}
	\begin{equation}
		\mu_i = (n-1)\cdot {\rm mean}(J_{i:}),
		\label{eqn:mu}
	\end{equation}
	\begin{equation}
		s_i=\sqrt{(n-1)\cdot {\rm Var}([J_{i:};-J_{i:}])}.
		\label{eqn:sd}
	\end{equation}
	\label{eqn:local_energy}
\end{subequations}
where the semicolon $(;)$ is used to indicte concatenation of the two vectors.

\subsection{Hyperparameter determination}

\cref{fig:selection} (b)  illustrates how to determine   the three hyperparameters of $I_{0min}$, $I_{0max}$, and $n_{rnd}$ using the local energy distribution.
As the three hyperparameters are common for all the spins, as shown in \cref{eqn:SSA}, the mean of the local energy distributions are considered.
In other words, a normal distribution $N({\rm mean}(\left|\mu_i\right|), {\rm mean}(s_i^2))$ is used to control $I_0$ that commonly controls for all the spins.

First, the magnitude of noise signals $n_{rnd}$ is determined as follows:
\begin{equation}
	n_{rnd} = 0.6745\cdot {\rm mean}(s_i).
	\label{eqn:nrnd}
\end{equation}
The reason to select 0.6745 is that the point where the area from the mean of the normal distribution equals 50\% is at $\pm$0.6745$\cdot \sigma$ from the mean.
It means that the local energy distribution can be equally divided into two area by $n_{rnd}$.
Second, the minimum pseudo inverse temperature $I_{0min}$ is determined as follows:
\begin{equation}
	I_{0min} = 0.01\cdot{\rm max}(s_i)+{\rm min}(\left|\mu_i\right|).
	\label{eqn:I0min}
\end{equation}
$I_{0min}$ is set as close to ${\rm min}(\left|\mu_i\right|)$ as possible to facilitate the flipping of the spin states.
Third, the maximum pseudo inverse temperature $I_{0max}$ is determined as follows:
\begin{equation}
	I_{0max} = 2\cdot{\rm max}(s_i)+{\rm min}(\left|\mu_i\right|).
	\label{eqn:I0max}
\end{equation}
It means that a cumulative probability within  a distance of $2\cdot {\rm max}(s_i)$ from the mean is approximately 95\%.
The reasoning behind the selection of max/min functions originates from the local energy distribution. 
Each local distribution at its spin has different $s_i$ and $\mu_i$, varying from small to large values. 
The usage of the max and min functions ensures that $I_0$ can encompass all local energy distributions by changing $I_{0min}$ to $I_{0max}$ during the annealing process.

\begin{table}
	\centering
	\caption{Summary of the determined hyperparameters for SSA.}
	\begin{tabular}{c|c}
		\hline
		$\mu_i$ & $(n-1)\cdot {\rm mean}(J_{i:})$ \\
		\hline
		$s_i$ & $\sqrt{(n-1)\cdot {\rm Var}([J_{i:};-J_{i:}])}$ \\
		\hline
		$n_{rnd}$ & $0.6745\cdot {\rm mean}(s_i)$ \\
		\hline
		$I_{0min}$ & $0.01\cdot{\rm max}(s_i)+{\rm min}(\left|\mu_i\right|)$ \\
		\hline
		$I_{0max}$ & $2\cdot {\rm max}(s_i)+{\rm min}(\left|\mu_i\right|)$ \\
		\hline
		$\beta$ & $\Bigl(\frac{I_{0min}}{I_{0max}}\Bigr)^{(\frac{1}{cycle-1})}$ \\
		\hline
	\end{tabular}
	\label{tb:SSA}
\end{table}

\cref{tb:SSA} summarizes the determined hyperparameters for SSA.
$\beta$ is determined by $(I_{0min}/I_{0max})^{(\frac{1}{cycle-1})}$ as $I_0$ is updated by $I_0(t+1)=I_0(t)/\beta$, as shown in  \cref{fig:hyperparameter}.
Note that $cycle$ is the total number of cycles during the annealing process.
The determined hyperparameters will be evaluated in \cref{sec:evaluation}.

\section{SSA WITH UNIQUE NOISE MAGNITUDE (SSAU)}
\label{sec:individual}

\begin{figure}[t]
	\centering
	\includegraphics[width=0.8\linewidth]{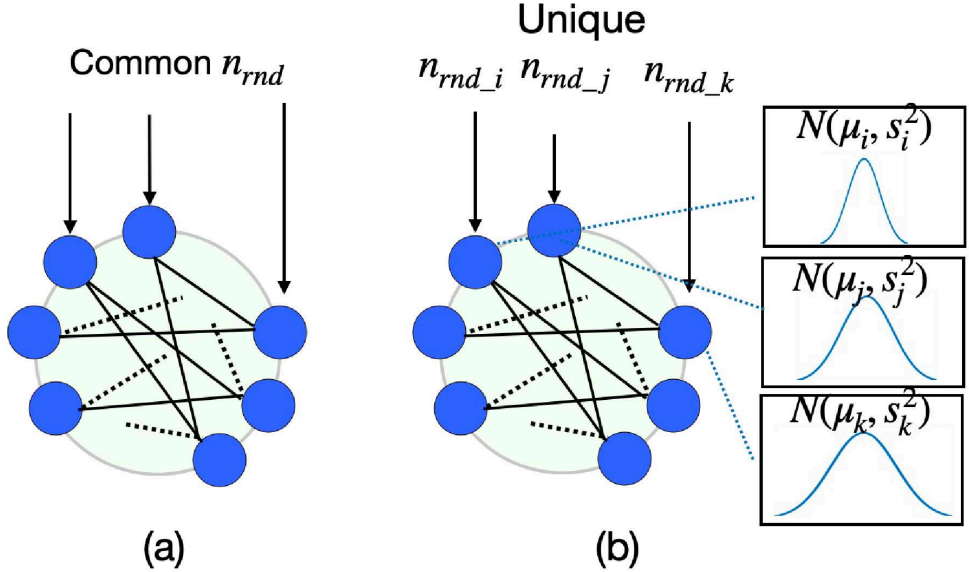}
	\caption{Magnitude of random signals: (a) SSA and (b) SSAU. In SSAU, a unique noise magnitude $n_{rnd\_i}$ is applied for each spin depending on its local energy distribution.}
	\label{fig:SSAU}
\end{figure}

In SSA, the magnitude of the noise signals $n_{rnd}$ is common for all spins, as shown in \cref{fig:SSAU} (a).
However, each spin has a unique local energy distribution that is different from others. 
To properly control the noise magnitude for each spin, SSA with a unique noise magnitude (SSAU) is introduced as an extension of SSA. 
This approach aims to better control the noise level for each spin and improve the search performance.

\cref{fig:SSAU} (b) shows SSAU with a unique noise magnitude $n_{rnd_i}$ for the $i$-th spin. 
The noise magnitude for SSA in \cref{eqn:nrnd} is updated for SSAU as follows:
\begin{equation}
	n_{rnd_i} = 0.6745\cdot s_i,
	\label{eqn:nrndi_u}
\end{equation}
where $s_i$ is the standard deviation of the local energy distribution at $i$-th spin in \cref{eqn:local_energy}. 
The other hyperparameters for SSAU are the same as that of SSA as shown in \cref{tb:SSA}.

Using $n_{rnd\_i}$, \cref{eqn:I_SSA} is replaced by the following equation:
\begin{equation}
	I_i(t+1) = h_i + \sum_j J_{ij}\cdot \sigma_j(t) + n_{rnd\_i}\cdot r_i(t).
	\label{eqn:I_SSAU}
\end{equation}
The other equations of \cref{eqn:SSA} are common for SSA and SSAU.

\section{EVALUATION}
\label{sec:evaluation}

\subsection{Simulation setup}

\begin{figure}[t]
	\centering
	\includegraphics[width=0.5\linewidth]{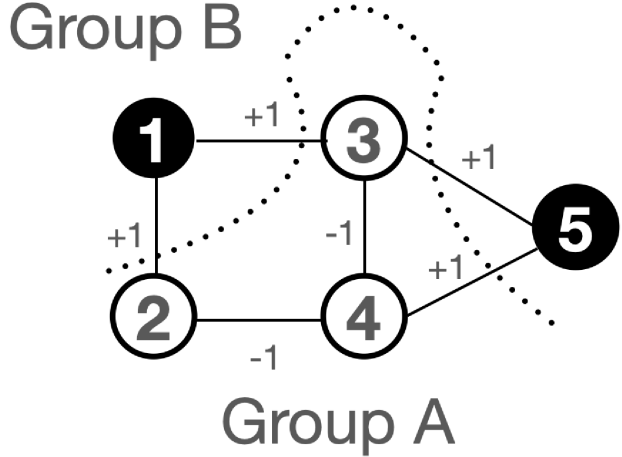}
	\caption{An example of a five-node MAX-CUT problem with edge weights of $-1$ and $+1$. The line cuts the edges to divide the graph into two groups while the sum of the edge weights is maximized.}
	\label{fig:maxcut}
\end{figure}

The proposed statistical methods for determining hyperparameters are evaluated on maximum-cut (MAX-CUT) problems, which are a typical class of combinatorial optimization problems \cite{max-cut}. 
An example of a five-node MAX-CUT problem with edge weights of $-1$ and $+1$ is shown in \cref{fig:maxcut}. 
The objective of MAX-CUT is to maximize the sum of edge weights by dividing the graph into two groups through a line cut.
The annealing process changes the spin states in an attempt to reach the global minimum energy described in \cref{eqn:Ising_SA}, where the maximum cut value is corresponding to the global minimum energy.
The black circle illustrates a spin state of '+1', while the white circle illustrates a spin state of '-1'.
In this example, the graph is divided into Group A (nodes 2, 3, and 4) and Group B (nodes 1 and 5), with a sum of edge weights equal to 4. 
The edge weights are represented using $J$ in the Hamiltonian shown in \cref{eqn:Ising_SA}.

\begin{table}[t]
	\centering
	\caption{Summary of  MAX-CUT problems used for evaluation.}
	\begin{tabular}{c||c|c|c|c}
		\hline
		Graph  & \# nodes & Structure & Weights ($J_{ij}$) & \# edges \\
		& & &  & $(J_{ij}\neq 0)$ \\
		\hline
		\hline
		G1 &800 & random  & $\{+1, 0\}$ &19176 \\
		\hline
		G6  & 800  & random & $\{+1,0, -1\}$ &19176 \\
		\hline
		G11  &800  & toroidal & $\{+1, 0, -1\}$  &1600 \\
		\hline
		G14  & 800  & planar & $\{+1, 0\}$ & 4694\\
		\hline
		G18  &  800 & planar & $\{+1, 0, -1\}$ &4694  \\
		\hline
		G22 &  2000 & random & $\{+1, 0\}$ &19990\\
		\hline
		G34  & 2000 & toroidal & $\{+1,0, -1\}$  &4000  \\
		\hline
		G38 & 2000 & planar &  $\{+1,0\}$ &11779\\
		\hline
		G39  &  2000 & planar & $\{+1, 0,-1\}$ &11778  \\
		\hline
		G47&  1000 & random  & $\{+1,0\}$ &9990 \\
		\hline
		G48 &  3000 & toroidal & $\{+1,0, -1\}$ &6000 \\
		\hline
		G54 & 1000 & random &  $\{+1,0\}$ &5916 \\
		\hline
		G55 & 5000 & random & $\{+1,0\}$  &12498  \\
		\hline
		G56  & 5000 & random & $\{+1,0, -1\}$  &12498   \\
		\hline
		G58 &  5000 & planar & $\{+1,0\}$ &29570 \\
		\hline
		K2000 & 2000 & full & $\{+1, -1\}$  &1999000  \\
		\hline
	\end{tabular}
	\label{tb:graph}
\end{table}

\cref{tb:graph} summarizes the benchmarks for the MAX-CUT problems that are used to evaluate the proposed method. 
The Gset includes the Gx graphs with different sizes, shapes, and weights \cite{G-set}, while K2000 is another benchmark that consists of fully-connected graphs with 2,000 nodes \cite{K2000}.
All simulation results are obtained using Python 3.9.6 on an Apple M1 Ultra with 128 GB of memory.
%

\subsection{Determined hyperparameters}

\begin{table*}[t]
	\centering
	\caption{Summary of  hyperparameters determined for SSA and SSAU.}
	\begin{tabular}{c||c|c||c|c|c|c}
		\hline
		Graph & $\left|\mu_i\right|$ & $s_i$  & \multicolumn{3}{c|}{SSA}  & SSAU \\ 
		\cline{4-7}
		&&  &  $n_{rnd}$ & $I_{0min}$ & $I_{0max}$ &   $n_{rnd\_i}$  \\
		\hline
		\hline
		G1 & [26.97, 66.92] & [5.19, 8.18] & 4.66 & 27.05 & 43.33 & [3.50, 5.52] \\
		G6 & [0.00, 28.96] & [5.19, 8.18] & 4.66 & 0.08 & 16.36 & [3.50, 5.52] \\
		G11 & [0.00, 3.99] & [1.99, 1.99] & 1.35 & 0.02 & 3.99 & [1.35, 1.35] \\
		G14 & [4.99, 131.84] & [2.23, 11.48] & 2.18 & 5.11 & 27.96 & [1.50, 7.74] \\
		G18 & [0.00, 17.98] & [2.23, 11.48] & 2.18 & 0.11 & 22.96 & [1.50, 7.74] \\
		G22 & [6.99, 36.98] & [2.64, 6.08] & 2.99 & 7.05 & 19.16 & [1.78, 4.10] \\
		G34 & [0.00, 3.99] & [1.99, 1.99] & 1.35 & 0.02 & 3.99 & [1.35, 1.35] \\
		G38 & [3.99, 248.88] & [1.99, 15.78] & 2.17 & 4.16 & 35.55 & [1.35, 10.64] \\
		G39 & [0.00, 42.98] & [1.99, 14.49] & 2.17 & 0.14 & 28.97 & [1.35, 9.77] \\
		G47 & [7.99, 33.97] & [2.83, 5.83] & 2.99 & 8.05 & 19.64 & [1.90, 3.93] \\
		G48 & [3.99, 3.99] & [1.99, 1.99] & 1.35 & 4.02 & 7.99 & [1.35, 1.35] \\
		G54 & [4.99, 135.86] & [2.23, 11.66] & 2.18 & 5.11 & 28.30 & [1.51, 7.86] \\
		G55 & [0.00, 14.99] & [0.00, 3.87] & 1.46 & 0.03 & 7.75 & [0.00, 2.61] \\
		G56 & [0.00, 9.99] & [0.00, 3.87] & 1.46 & 0.03 & 7.75 & [0.00, 2.61] \\
		G58 & [3.99, 560.88] & [1.99, 23.68] & 2.17 & 4.24 & 51.36 & [1.35, 15.97] \\
		K2000 & [0.99, 168.92] & [44.70, 44.70] & 30.15 & 1.45 & 90.40 & [30.15, 30.15] \\
		\hline
	\end{tabular}
	\label{tb:hyperparameter}
\end{table*}

\begin{figure}[t]
	\centering
	\includegraphics[width=0.5\linewidth]{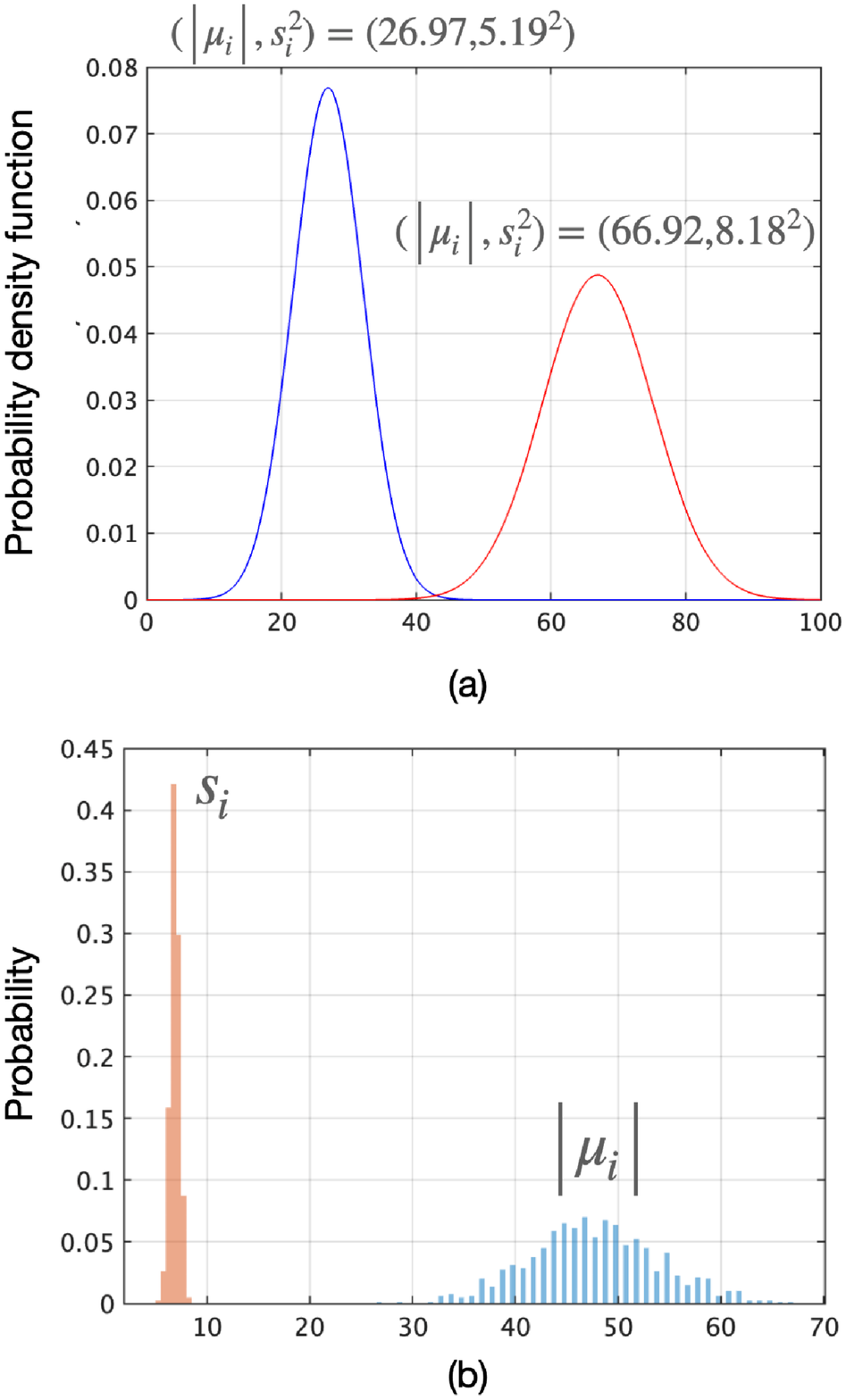}
	\caption{$\left|\mu_i\right|$ and $s_i$ for G1: (a) two normal distribution with the minimum and the maximum $\left|\mu_i\right|$ as an example and (b) the histogram of $\left|\mu_i\right|$ and $s_i$ for all 800 spins.}
	\label{fig:G1}
\end{figure}

\cref{tb:hyperparameter} summarizes the hyperparameters used for SSA and SSAU. 
The equations for calculating these hyperparameters are provided in \cref{tb:SSA} and \cref{eqn:nrndi_u}.
To illustrate how these values are determined, let us consider the example of G1. 
The matrix $J$ of G1 is an 800$\times$800 matrix. 
It consists of edge weights either `+1' or `0', with the count of `+1' edge weights being 19,176. 
The total number of edges in the matrix amounts to 319,600.
%
First, $\left|\mu_i\right|$ and $s_i$ are calculated for each of the 800 vectors in $J$, which yields a unique set of values for each spin. 
As an example, two normal distributions with the minimum and maximum values  of $\mu$ are shown in \cref{fig:G1} (a). 
The range of $\left|\mu_i\right|$ is from 26.97 to 66.92, and the range of $s_i$ is from 5.19 to 8.18, as shown in \cref{fig:G1} (b).

Next, $\left|\mu_i\right|$ and $s_i$  are used to calculate the hyperparameters of $n_{rnd}$, $I_{0min}$, and $I_{0max}$. 
In the case of SSAU, each spin has a unique $n_{rnd\_i}$ value, which corresponds to its local energy distribution. 
The range of $n_{rnd\_i}$ is from 3.50 to 5.52.

\begin{figure}[t]
	\centering
	\includegraphics[width=0.6\linewidth]{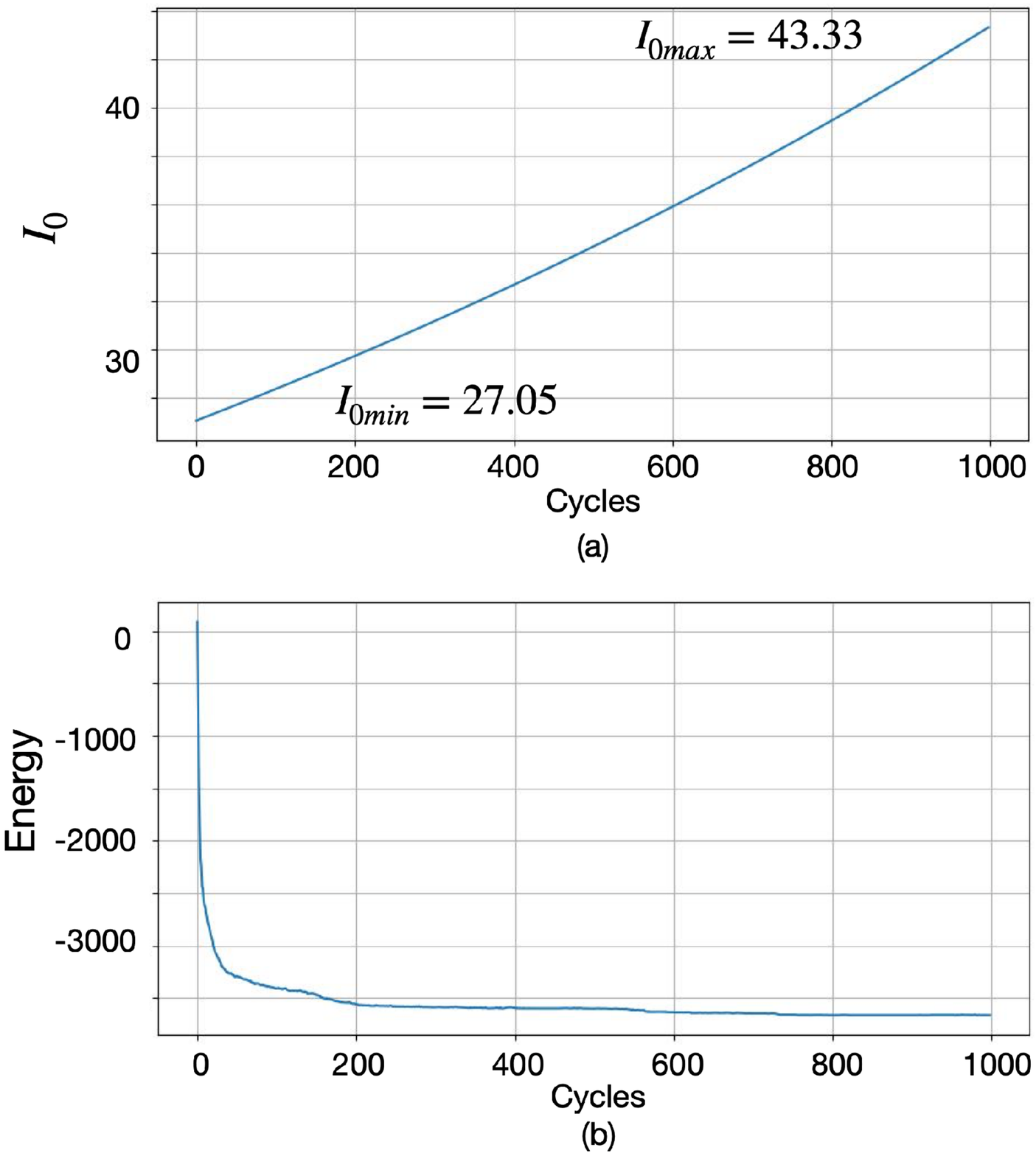}
	\caption{Simulation example of SSA with determined hyperparameters for G1: (a) $I_0$ vs. cycles and (b) energy vs. cycles. The energy decreases to the global minimum energy as $I_0$ is increased from $I_{0min}$ and $I_{0max}$.}
	\label{fig:sim_example}
\end{figure}

\begin{table*}[t]
	\centering
	\caption{Comparisons of performance on the MAX-CUT benchmarks for 1,000 cycles.}
	\begin{tabular}{c|c||c|c|c|c|c|c||c|c|c}
		\hline
		Graph  & Best known & \multicolumn{3}{c|}{Mean of cut values} & \multicolumn{3}{c||}{Standard deviation of cut values} & \multicolumn{3}{c}{Ratio of mean cut value}\\
		\cline{3-11}
		&cut value& SA \cite{SA_max-cut} & SSA & SSAU & SA & SSA & SSAU & SSA  & SSAU  & SSAU  \\
		& & & & & & & & vs. SA & vs. SA & vs. SSA \\
		\hline
		\hline
		G1 & 800 & 10754.52 & 11427.05 & 11428.13 & 47.63 & 32.83 & 33.28 & 6.25\% & 6.26\% & 0.01\% \\
		G6 & 800 & 1276.96 & 2159.59 & 2160.63 & 45.56& 12.29 & 12.44 & 69.12\% & 69.20\% & 0.05\% \\
		G11 & 800 & 334.98 & 549.60 & 549.47 & 13.16 & 4.12 & 4.10& 64.07\% & 64.03\% & -0.02\% \\
		G14 & 800 & 2803.34 & 3009.71 & 3013.20 & 15.55 & 8.13 & 7.70 & 7.36\% & 7.49\% & 0.12\% \\
		G18 & 800 & 590.68 & 972.49 & 974.72 & 25.60 & 8.34 & 7.56 & 64.64\% & 65.02\% & 0.23\% \\
		G22 & 2000 & 11161.3 & 13099.78 & 13102.26 & 50.47& 31.63 & 31.34 & 17.37\% & 17.39\% & 0.02\% \\
		G34 & 2000 & 469.26 & 1346.64 & 1346.67 & 27.97 & 6.34 & 6.39 & 186.97\% & 186.98\% & 0.00\% \\
		G38 & 2000 & 6642.68 & 7546.93 & 7554.4 & 31.09 & 15.82 & 15.71 & 13.61\% & 13.73\% & 0.10\% \\
		G39 & 2000 & 862.1 & 2352.47 & 2362.03 & 46.48 & 14.01 & 12.41 & 172.88\% & 173.99\% & 0.41\% \\
		G47 & 1000 & 5854.9 & 6536.24 & 6537.83 & 33.10 & 21.62 & 21.15 & 11.64\% & 11.66\% & 0.02\% \\
		G48 & 3000 & 3562.2 & 5724.25 & 5724.29 & 30.94 & 38.20 & 38.15 & 60.69\% & 60.70\% & 0.00\% \\
		G54 & 1000 & 3484.95 & 3780.36 & 3784.71 & 19.66 & 8.73 & 8.54 & 8.48\% & 8.60\% & 0.11\% \\
		G55 & 5000 & 6973.55 & 9994.42 & 10037.44 & 45.21 & 22.33 & 21.02 & 43.32\% & 43.94\% & 0.43\% \\
		G56 & 5000 & 702.21 & 3930.36 & 3947.27 & 55.74 & 14.72 & 12.08& 459.71\% & 462.12\% & 0.43\% \\
		G58 & 5000 & 15791.8 & 18930.60 & 18949.24 & 68.81 & 27.41 & 30.40 & 19.88\% & 19.99\% & 0.10\% \\
		K2000 & 2000 & 11267.73 & 32931.54 & 32932.38 & 566.15 & 117.59 & 117.71 & 192.26\% & 192.27\% & 0.00\% \\
		\hline
	\end{tabular}
	\label{tb:comparison}
\end{table*}

To verify the hyperparameters, SSA with G1 is simulated as an example. 
\cref{fig:sim_example} (a) illustrates the pseudo inverse-temperature transition  for 1,000 cycles. 
$I_0$ is  gradually increased  from $I_{0min}=27.05$ to $I_{0max}=43.33$. 
The value of $I_0$ is updated by $I_0(t+1)=I_0(t)/\beta$, where $\beta=0.99952$.
The energy, defined by \cref{eqn:Ising_SA}, decreases to the global minimum as $I_0$ increases. 
Once the simulation completes 1,000 cycles, the spin states $\sigma_i$ are extracted to compute the cut value using $J$.

\subsection{Comparisons}

\cref{tb:comparison} compares SSA and SSAU with a conventional SA in the 16 different MAX-CUT problems for 1,000 cycles.
Let us briefly explain the simulation conditions for SA, as detailed in \cite{SA_max-cut}. 
In this method, the temperature $T$ is gradually decreased by $\Delta_{IT}$ as $T \leftarrow 1/(1/T+\Delta_{IT})$ at each cycle.
The initial temperature is set to 1, and the final temperature is set to 1/1000.
The number of cycles is 1,000, which is the same as for SSA and SSAU.
During each cycle, a spin state is randomly flipped, and a new state is accepted if the new energy ($E_{new}$) is lower than the current energy ($E_{cur}$) or if it is higher with a probability of ${\rm exp}(-(E_{new}-E_{cur})/T)$.

The mean and the standard deviation of cut values are obtained by running 100 trials for each benchmark. 
When comparing SSA and SSAU with SA, the ratios of the mean cut values are positive in all benchmarks. 
In particular, the ratios are significantly positive in the case of graphs with \{+1, 0, -1\} weights. 
The reason is that SA takes more cycles to achieve better cut values in the case, where the mean cut values are significantly smaller than the best-known values.
The mean of normalized cut values is 65.0\% for SA.

\begin{table}[h]
	\centering
	\caption{Summary of best cut values for 1,000 cycles.}
	\begin{tabular}{c||c|c|c}
		\hline
		& SA & SSA & SSAU \\ \hline \hline
		G1   & 10859 & 11550& 11560 \\ \hline
		G6   & 1389 & 2178 & 2178 \\ \hline
		G11  & 364 & 564 & 564 \\ \hline
		G14  & 2836 & 3040 & 3040 \\ \hline
		G18  & 647 & 990 & 992 \\ \hline
		G22  & 11325 & 13214 & 13221 \\ \hline
		G34  & 536 & 1368 & 1368 \\ \hline
		G38  & 6715 & 7603 & 7606 \\ \hline
		G39  & 993 & 2399 & 2400 \\ \hline
		G47  & 5947 & 6647 & 6615 \\ \hline
		G48  & 3642 & 5934 & 5904 \\ \hline
		G54  & 3530 & 3820 & 3815 \\ \hline
		G55  & 7088 & 10073 & 10121 \\ \hline
		G56  & 831 & 3982 & 3996 \\ \hline
		G58  & 15919 & 19014 & 19033 \\ \hline
		K2000 & 12413 & 33235 & 33225 \\ \hline
	\end{tabular}
	\label{tb:best}
\end{table}

\begin{figure}[t]
	\centering
	\includegraphics[width=0.6\linewidth]{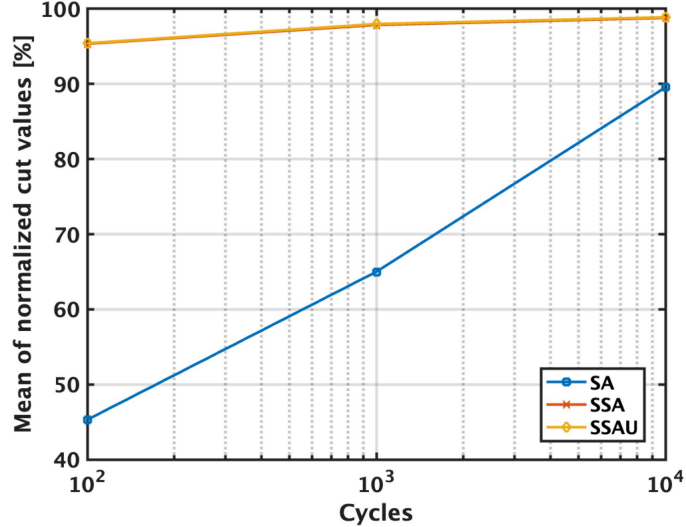}
	\caption{The mean of normalized cut values versus cycles for 16 benchmarks. It shows that both SSA and SSAU rapidly attain an average of 95.4\% of the best-known normalized value within just 100 cycles. In comparison, SA only achieves 48.2\%. }
	\label{fig:value_cycle}
\end{figure}

\begin{table}[t]
	\centering
	\caption{Time complexity of hyperparameter search in SSA.}
	\begin{tabular}{c|c|c}
		\hline
		& Conventional  \cite{SSA,ICECS2022} & This work \\
		\hline
		Time complexity & $\mathcal{O}(n^3)$ & $\mathcal{O}(1)$ \\
		\hline
	\end{tabular}
	\label{tb:order}
\end{table}

\begin{table*}[t]
	\centering
	\caption{Comparisons of search and annealing time with normalized mean cut values for 1000 cycles on all 16 benchmarks using SSA.}
	\label{tab:my-table}
	\begin{tabular}{c||c|c|c|c||c|c|c}
		\hline
		& \multicolumn{3}{c|}{Conventional} & This work &  \multicolumn{3}{c}{Normalized mean cut value}\\
		\cline{2-8}
		& Search time [s] & Annealing time [s] & Total time [s] & Annealing time  & Conventional & This work & Ratio\\
		& & & &  (total time)  [s]  & & & \\
		\hline
		\hline
		G1 & 310.61 & 0.28 & 310.88 & 0.40 & 98.58\% & 98.31\% & -0.28\% \\
		G6 & 286.76 & 0.26 & 287.02  & 0.28 & 89.51\% & 99.15\% & 10.78\% \\
		G11 & 313.34 & 0.28 & 313.62  & 0.40 & 88.1\%4 & 97.45\% & 10.56\% \\
		G14 & 312.60 & 0.25 & 312.85 & 0.39 & 97.44\% & 98.23\% & 0.81\% \\
		G18 & 316.05 & 0.28 & 316.32 & 0.29 & 87.90\% & 98.03\% & 11.53\% \\
		G22 & 1523.29 & 1.45 & 1524.74 & 1.45 & 97.4\%2 & 98.06\% & 0.66\% \\
		G34 & 1494.86 & 1.45 & 1496.30  & 1.80 & 94.66\% & 97.30\% & 2.79\% \\
		G38 & 1483.77 & 1.45 & 1485.21  & 1.83 & 96.95\% & 98.17\% & 1.26\% \\
		G39 & 1473.18 & 1.45 & 1474.63  & 1.83 & 87.2\%3 & 97.69\% & 12.00\% \\
		G47 & 433.96 & 0.41 & 434.37  & 0.61 & 96.92\% & 98.18\% & 1.31\% \\
		G48 & 3250.20 & 3.26 & 3253.46 & 4.19 & 88.33\% & 95.40 \%& 8.01\% \\
		G54 & 418.73 & 0.40 & 419.12 & 0.61 & 97.81\% & 98.14\% & 0.34\% \\
		G55 & 8812.68 & 8.96 & 8821.64  & 10.71 & 94.68\% & 97.04\% & 2.49\% \\
		G56 & 8869.97 & 8.80 & 8878.77  & 9.30 & 94.24\% & 97.84\% & 3.83\% \\
		G58 & 8794.97 & 8.82 & 8803.79  & 9.24 & 97.57\% & 98.12\% & 0.57\% \\
		K2000 & 1451.41 & 1.44 & 1452.85 & 1.52 & 98.24\% & 98.78\% & 0.56\% \\ 
		\hline
	\end{tabular}
	\label{tb:time}
\end{table*}

When comparing SSA and SSAU, the mean cut values of SSAU are larger than that of SSA in most of the benchmarks. 
The ratios are relatively large in the case of graphs with \{+1, 0\} weights. 
The standard deviation of the cut values is similar for both methods, with these values significantly smaller than that of SA. 
On average, across all benchmarks for 1,000 cycles, the mean cut values of SSA and SSAU are 97.9\% and 98.0\% of the best-known cut values, respectively.
In terms of computation cost, SSAU takes almost the same simulation time as SSA. 
For instance, the simulation time on K2000 for 1,000 cycles is 1.43 seconds for both SSA and SSAU.
The best cut values are summarized in \cref{tb:best}. 
It shows that SSA and SSAU achieve 99.2\% of the best-known cut values while SA only attains 67.7\%.

\cref{fig:value_cycle} displays the mean of normalized cut values in relation to the number of cycles for 16 benchmarks. 
Both SSA and SSAU quickly achieve an average of 95.4\% of the normalized best-known value within just 100 cycles, while SA only reaches 48.2\%.
For 10,000 cycles, SA, SSA, and SSAU achieve 90.0\%, 98.8\%, and 98.9\% respectively.
Compared to related works on K2000, the mean cut value of SSAU is 32,932, which surpasses the 32,458 and 32,768 values from \cite{K2000} and \cite{SB}, respectively.

\section{DISCUSSION}
\label{sec:discussion}

\subsection{Comparison with hyperparameter search}

The comparison result between SA and SSA for G11 was previously presented in \cite{SSA}. 
SSA achieved superior mean cut values compared to SA and QA, even though the previous study employed hyperparameter searching.
\cref{tb:order} provides a comparison of the time complexity of hyperparameter searching for SSA. 
In the conventional method \cite{SSA,ICECS2022}, the hyperparameters of $n_{rnd}$, $I_{0min}$, and $I_{0max}$ were selected through searching, leading to a time complexity of $\mathcal{O}(n^3)$.
In contrast, the proposed method determines these hyperparameters statistically without the need for searching, thereby resulting in a time complexity of $\mathcal{O}(1)$.

Let us discuss the impact of determined hyperparameters in SSA. 
As summarized in \cref{tb:order}, the conventional method necessitates the search for hyperparameters prior to simulated annealing.
\cref{tb:time} compares the search and annealing time, along with the normalized mean cut values  for the all 16 benchmarks using SSA, where the annealing takes 1,000 cycles.
In the conventional method, the hyperparameters of $I_{0min}$, $I_{0max}$, and $n_{rnd}$ are randomly searched for across 1,000 trials, with each value ranging from 0 to 1000 with a  condition of $I_{0min} \geq I_{0min}$.
Based on these search results, the best hyperparameters are selected to evaluate the mean cut values.
The outcome indicates that the proposed method, which excludes the hyperparameter search, yields results that are superior to the conventional method in most benchmarks.
Moreover, the proposed method completely eliminates the time-consuming process of hyperparameter searching.

\begin{figure*}[t]
	\centering
	\includegraphics[width=1.0\linewidth]{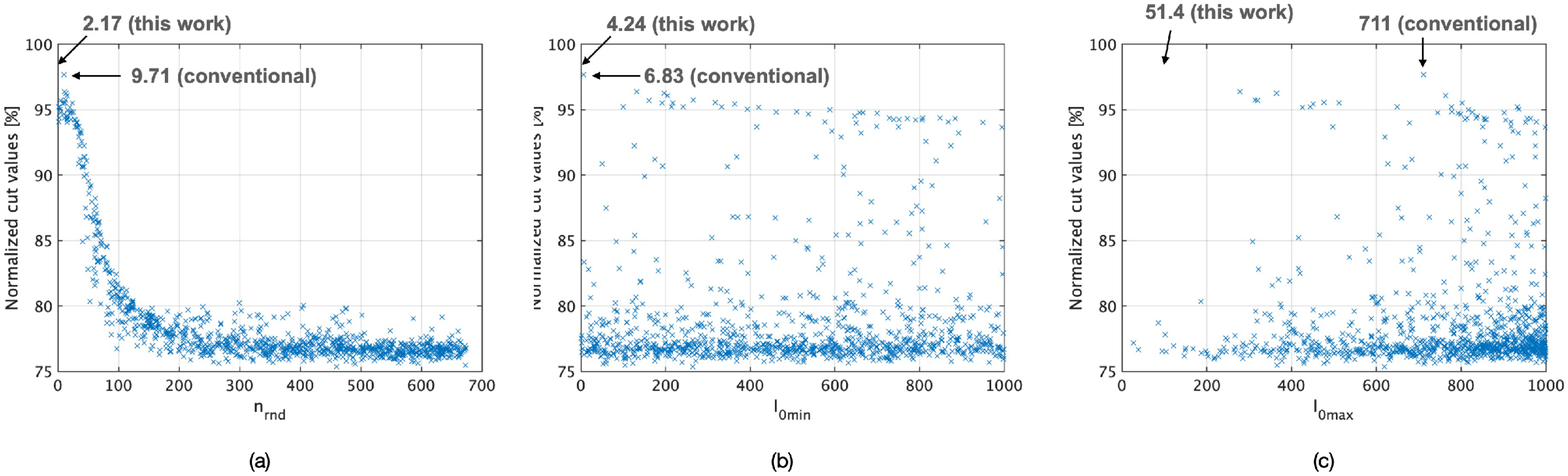}
	\caption{1,000 normalized cut values using random search for 1,000 cycles on G58 are presented as: (a) $n_{rnd}$, (b) $I_{0min}$, and (c) $I_{0max}$. The conventional method achieves a normalized mean cut value of 97.6\% with the best hyperparameters searched, while the proposed method achieves 98.2\% with the determined hyperparameters. Both sets of parameters are plotted in the figures.}
	\label{fig:parameter}
\end{figure*}

\cref{fig:parameter} illustrates the normalized cut values using random search with 1,000 cycles on G58. 
In G58, the proposed method achieves a normalized mean cut value of 98.2\%, which is superior to the 97.6\% of the conventional method.
Both the best hyperparameters identified through searching and the determined hyperparameters are plotted.
The plotted values for $n_{rnd}$ and $I_{0min}$ are similar between the conventional and proposed methods; however, the values for $I_{0max}$ differ significantly, potentially leading to performance loss.
As the search process iterates 1,000 times, more iterations might yield better hyperparameters, but this would also increase the search time.

\subsection{Hyperparameter selection}

Let us discuss the effect of using 0.01 for $I_{0min}$ in \cref{eqn:I0min}. 
In this study, the constant 0.01 is utilized to position $I_{0min}$ close to ${\rm min}(\left|\mu_i\right|)$. 
This results in a mean normalized cut value of 97.9\% on average in SSA.
When 0.01 is changed to $10^{-6}$ or $0.05$, the respective averages become 97.6\% and 97.8\%. 
Hence, the constant for $I_{0min}$ does not significantly impact the results.
Note that a value of 0 is not permissible.
The range of this value is associated with the equation $\beta = \Bigl(\frac{I_{0min}}{I_{0max}}\Bigr)^{(\frac{1}{cycle-1})}$ as shown in Table 1. 
If the value is 0, $I_{0min}$ becomes 0 in several benchmarks, especially when ${\rm min}(\left|\mu_i\right|)$ is 0.
This makes it impossible to determine $\beta$. 
Therefore, $I_{0min}$ is set to a value as low as $10^{-6}$.

Additionally, the number of standard deviations set to 2 for $I_{0max}$ in \cref{eqn:I0max} is examined. 
To evaluate the impact of this standard deviation number, it is altered to 3 and 4. 
As a result, the mean normalized cut value averages 97.8\% in SSA for both cases. 
Therefore, the constant for $I_{0max}$ also does not have a significant effect on the results.

\begin{figure}[t]
	\centering
	\includegraphics[width=0.6\linewidth]{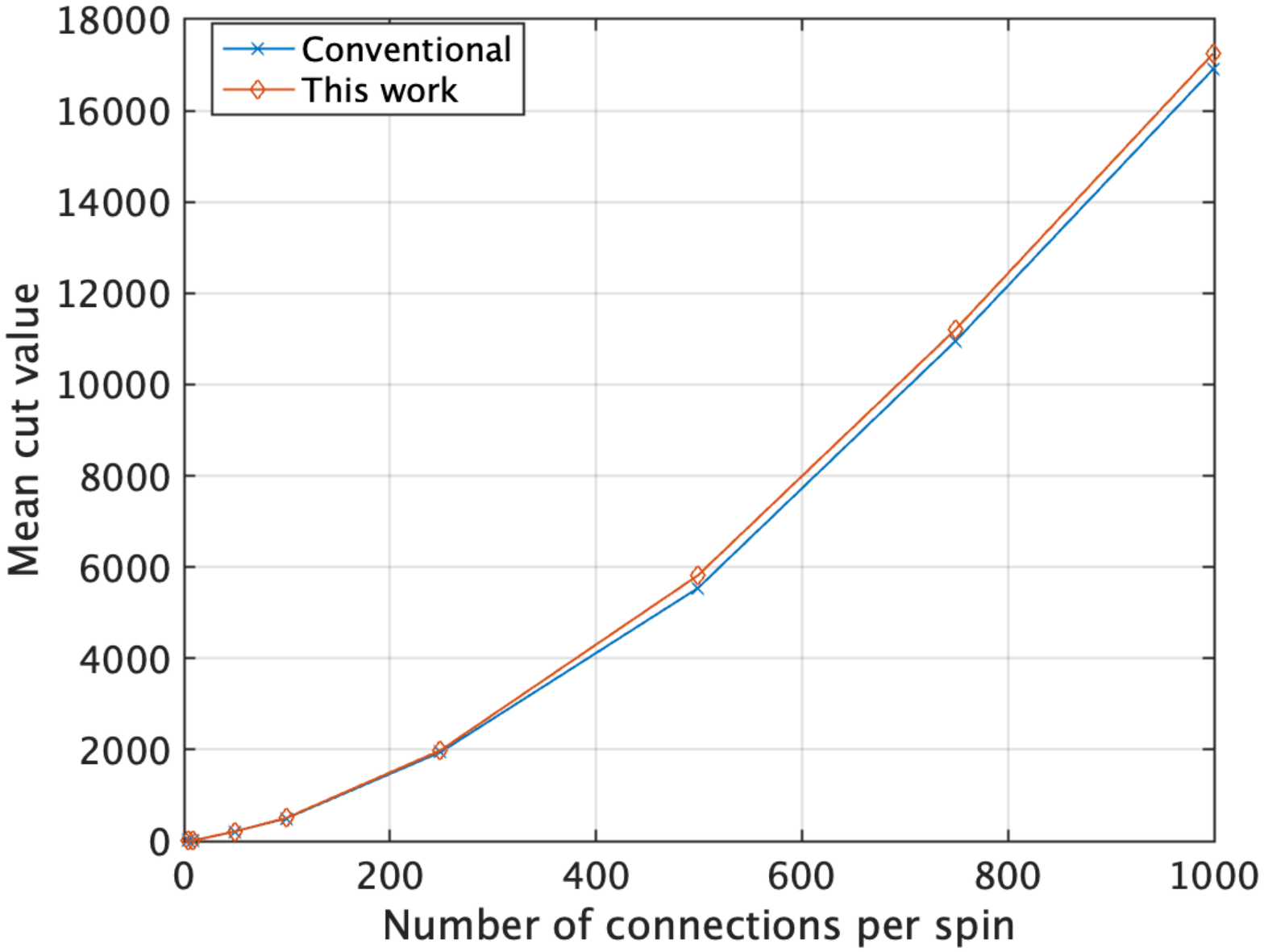}
	\caption{Mean cut values versus the number of connections per spins in fully-connected benchmark graphs with weights of +1 or -1. }
	\label{fig:CLT}
\end{figure}

There might be a constraint of the proposed determined hyperparameters based on CLT when each spin contains a few connections.
To assess the performance of the proposed method based on the number of connections, new benchmarks are created using K2000 as a reference. K2000 is fully connected with weights of +1 or -1, and the ratio of +1 to -1 weights is 50\%.
These benchmarks encompass spin counts ranging from 5 to 1000, with each spin being fully connected and assigned weights of either +1 or -1.
The mean cut values, taken from 100 trials, are contrasted between the conventional search and the proposed method in \cref{fig:CLT}.
The conventional method searches  the best hyperparameters for 1,000 trials.
The findings indicate that the proposed method outperforms the conventional approach when there is a larger number of connections.
However, with a smaller number of connections, the assumptions underpinning the Central Limit Theorem (CLT) might not be met, even though the mean cut values remain nearly identical.
A possible explanation is that annealing is effective with a smaller number of spins, even in the absence of precise hyperparameters.

\section{CONCLUSION}
\label{sec:conclusion}

This paper proposes the local energy distribution based  hyperparameter determination in SSA, which is a faster solving method for combinatorial optimization problems than SA. 
The method is based on the local energy distributions of spins and utilizes the CLT-based normal distribution for hyperparameter determination, significantly reducing the time complexity for hyperparameter search. 
Additionally, using the local energy distributions, the unique magnitude for the random signal at each spin has been presented to further improve SSA. 
The proposed method is evaluated in the MAX-CUT problems, where it achieves high accuracy while reducing time costs compared to conventional SSA with hyperparameter search.

Future research directions include applying the proposed method to other optimization problems, exploring the impact of hyperparameters on the SSA performance, developing more efficient hardware.

\section*{Acknowledgment}

This work was supported in part by JST CREST Grant Number JPMJCR19K3, and JSPS KAKENHI Grant Number JP21H03404.

\bibliographystyle{unsrtnat}
\bibliography{OJSP}

\end{document}